\pdfoutput=1

\documentclass[11pt]{article}

\usepackage[]{ACL2023}
\usepackage{xcolor}
\definecolor{darkgreen}{rgb}{0.0, 0.5, 0.0}
\definecolor{lightblue}{RGB}{173,216,230}
\definecolor{lightred}{RGB}{255,182,193}
\definecolor{lightgreen}{RGB}{173,255,47}
\definecolor{lightyellow}{RGB}{255,255,204}
\definecolor{violet}{RGB}{90, 19, 242}

\usepackage{amsmath,amsfonts,bm}

\def\eqref#1{equation~\ref{#1}}

\def\1{\bm{1}}

\DeclareMathAlphabet{\mathsfit}{\encodingdefault}{\sfdefault}{m}{sl}
\SetMathAlphabet{\mathsfit}{bold}{\encodingdefault}{\sfdefault}{bx}{n}

\usepackage{times}
\usepackage{latexsym}
\usepackage{graphicx}
\usepackage{subcaption}
\usepackage{hyperref}
\usepackage{wrapfig}
\usepackage{url}
\usepackage{graphicx}
\usepackage{multirow}
\usepackage{hyperref}
\usepackage{booktabs}
\usepackage{longtable}
\usepackage{longtable}
\usepackage{tabularx}
\usepackage{booktabs}
\usepackage{siunitx}
\usepackage{hyperref}
\usepackage{enumitem}
\usepackage{amsmath}
\usepackage{xcolor}
\usepackage{tcolorbox}
\usepackage{soul}
\usepackage{makecell}
\usepackage{multirow}
\usepackage{booktabs}
\usepackage{graphicx}
\usepackage{amsmath}
\usepackage{amssymb}
\usepackage{placeins}
\usepackage[titletoc]{appendix}
\usepackage{amssymb}
\usepackage{xcolor}
\usepackage{algorithm}
\tcbset{mybox/.style={colback=blue!5!white,colframe=blue!75!black,fonttitle=\bfseries,title=#1}}

\usepackage{algpseudocode}
\usepackage[T1]{fontenc}

\usepackage[utf8]{inputenc}

\usepackage{microtype}

\usepackage{inconsolata}

\title{Attn-GS: Attention-Guided Context Compression for Efficient Personalized LLMs }

\author{
  Shenglai Zeng$^1$\thanks{Work done during his internship at Amazon.}\thanks{zengshe1@msu.edu}, Tianqi Zheng$^2$, Chuan Tian$^2$, Dante Everaert$^2$, Yau-Shian Wang$^2$, Yupin Huang$^2$,\\ 
  \textbf{Michael J. Morais$^2$, Rohit Patki$^2$,Jinjin Tian$^2$, Xinnan Dai$^1$, Kai Guo$^1$,} \\
  \textbf{Monica Xiao Cheng$^2$, Hui Liu$^1$} \\
  \addlinespace[0.2cm] 
  $^1$Michigan State University \quad $^2$Amazon.com \\
   \\
  }

\begin{document}
\maketitle
\newtheorem{definition}{Definition}
\begin{abstract}
\label{abstract}
\vspace{-0.1cm}
Personalizing large language models (LLMs) to individual users requires incorporating extensive interaction histories and profiles, but input token constraints make this impractical due to high inference latency and API costs. Existing approaches rely on heuristic methods such as selecting recent interactions or prompting summarization models to compress user profiles. However, these methods treat context as a monolithic whole and fail to consider how LLMs internally process and prioritize different profile components. We investigate whether LLMs' attention patterns can effectively identify important personalization signals for intelligent context compression. Through preliminary studies on representative personalization tasks, we discover that (a) LLMs' attention patterns naturally reveal important signals, and (b) fine-tuning enhances LLMs' ability to distinguish between relevant and irrelevant information. Based on these insights, we propose \textbf{Attn-GS}, an attention-guided context compression framework that leverages attention feedback from a  marking model to mark important personalization sentences, then guides a compression model to generate task-relevant, high-quality compressed user contexts. Extensive experiments demonstrate that Attn-GS significantly outperforms various baselines across different tasks, token limits, and settings, achieving performance close to using full context while reducing token usage by 50 $\times$.

\end{abstract}
\vspace{-0.5cm}
\section{Introduction}
\label{Intro}
\begin{figure}[t]
    \centering
    \includegraphics[width=1\linewidth]{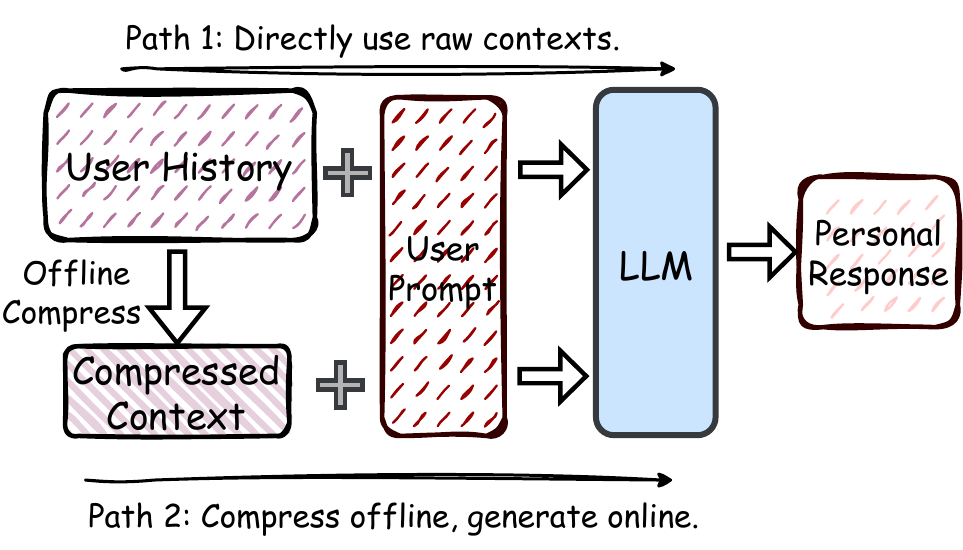}
    
    \caption{Personalized LLMs.}
    \vspace{-0.6cm}
    \label{fig:intro}
    
\end{figure}

\vspace{-0.1cm}
Large language models have been widely adopted across various applications, serving billions of users worldwide. In many applications such as query autocompletion~\cite{zhou2024cognitive,everaert2024amazonqac}, customer service~\cite{agarwal2022descriptive}, and content-based recommendation agents\cite{yang2023palr,zhang2024generative}, it is essential to adapt general LLMs into personalized models that can \textit{generate responses tailored to the unique needs and preferences of individual users}. The general personalization task can be formalized as shown in Figure~\ref{fig:intro}. Based on user profiles and interaction histories, the model is expected to understand user preferences and provide personalized responses to given tasks that align with each user's specific needs.
However, a fundamental challenge in personalization tasks is the constraint imposed by input token limits~\cite{zhu2025recommender,shi2025personax}. As users accumulate extensive interaction histories, directly inputting all contextual information into LLMs for personalization (Path 1 in Figure~\ref{fig:intro}) can result in high inference latency and API costs, making it impractical for real-world deployment\cite{zhu2025recommender,everaert2024amazonqac,10.1145/3711896.3736563}. Moreover, not all signals are relevant to specific tasks, and naively incorporating more signals into long contexts may not always yield benefits~\cite{lyu2023llm}. Therefore, as shown in the path 2 in Figure~\ref{fig:intro}, compressing extensive user histories and profiles offline into a condensed context within token limits while maintaining acceptable downstream performance becomes essential. Consequently,  it is crucial to identify which personalization signals are truly important for the task and intelligently select or compress these signals to satisfy context length requirements.

Current solutions primarily rely on heuristic or naive methods to address context limitation problems, such as selecting recent interactions~\cite{everaert2024amazonqac} or prompting summarization LLMs~\cite{liu2024once,zhang2024generative} to compress user profiles. However, these approaches treat the context as a monolithic whole and fail to consider how LLMs internally process and prioritize different profile components. Since  LLMs naturally exhibit selective attention~\cite{he2025attention,olsson2022context,gould2023successor} when processing input sequences—focusing more on relevant information while down-weighting less pertinent details—understanding these internal mechanisms could reveal which personalization signals the model \textbf{actually utilizes for generating responses}. Meanwhile, it is evident from other domains, such as evidence-based question answering, has demonstrated the crucial role of LLMs' internal representations~\cite{zeng2025towards} and attention patterns~\cite{liu2025selfelicit}, successfully leveraging these mechanisms to enhance performance. Inspired from the insights
, we investigate whether LLMs' attention behaviors can effectively identify important personalization signals and whether attention patterns can enable novel context compression strategies.
 
To answer these questions, we first conduct preliminary studies to examine attention weight distributions on representative personalization tasks in Section~\ref{Sec: Preliminaries}. We find that (a) LLMs' attention patterns naturally unveil important signals, and (b) fine-tuning enables LLMs to better distinguish between important and unimportant signals. Based on these findings, we develop an attention-guided context compression framework, \textit{Attn-GS}   (Section~\ref{Sec:Method}), to generate high-quality, task-relevant compressed profiles for downstream use. Specifically, we first utilize attention feedback from a small white-box marking model to identify important personalization sentences, then guide a compression model with these attention-marked contexts to generate task-relevant, high-quality compressed user contexts. Experiments (Section~\ref{Sec:experiment}) demonstrate that Attn-GS generates superior compressed profiles that outperform various baselines across different tasks, token limits and scenarios (inference-only and training-inference settings), achieving performance close to using full context while reducing token usage by 50$\times$.

\section{Related Work}\label{Related works}

\subsection{Personalized LLMs} 
LLM personalization refers to adapting Large Language Models to individual user preferences and contexts to deliver tailored responses. This paradigm has great potential across applications including education~\cite{wang2024large}, healthcare~\cite{yu2024health}, recommendation systems~\cite{yang2023palr}, and search~\cite{zhou2024cognitive}. Researchers employ various methods such as Retrieval-Augmented Generation (RAG)~\cite{zhao2024recommender,rajput2023recommender}, prompting~\cite{jiang2023evaluating,serapio2023personality}, and fine-tuning~\cite{li2021survey,zeng2025towards}. However, a key challenge lies in handling extensive user contexts: directly inputting them incurs prohibitive inference latency and API costs. Current approaches rely on either rule-based heuristics (e.g., using recent interactions or single signal types~\cite{dai2023uncovering,liu2023chatgpt,everaert2024amazonqac}) or LLM-based methods leveraging summarization~\cite{liu2024once,zhang2024generative} and self-reflection~\cite{zhang2024agentcf,wang2025user} to condense contexts. Yet significant gaps remain in understanding how LLMs internally process and prioritize different profile components.

\subsection{Utilization of LLMs' internal signals}
Recently, an emerging research direction focuses on utilizing LLMs' internal signals to understand how LLMs process contexts~\cite{halawi2023overthinking,chen2024context,li2024snapkv}. In question answering, researchers have shown that LLMs' internal representations~\cite{zeng2025towards2} can identify high-level concepts in RAG systems, such as context helpfulness, and can be controlled~\cite{zeng2025towards} to enhance RAG robustness. Moreover, \citet{liu2025selfelicit} demonstrate that LLMs possess ``evidence-seeking layers'' that identify evidence in QA tasks, and that attention scores can improve extractive QA performance. However, existing work predominantly focuses on QA tasks that extract explicit answers from context, while personalization requires synthesizing diverse user signals to generate tailored responses—a fundamentally different mechanism. Whether LLMs' internal representations can identify important personalization signals and facilitate context compression remains unexplored. We extend this exploration to the personalization domain.

\section{Preliminary Studies}
\label{Sec: Preliminaries}

In this section, we conduct preliminary studies on the inherent abilities of LLMs' attention mechanisms to distinguish between important and unimportant personalization signals. Specifically, we examine the distribution of LLMs' attention scores across different types of context signals to explore whether they can effectively differentiate important from unimportant signals. We first introduce the problem description and notations in Section~\ref{Sec:Description}, followed by our preliminary findings in Section~\ref{Sec:Findings}.

\subsection{Problem Description \& Notations}
\label{Sec:Description}
In this subsection, we introduce the problem setting and notations used in our paper. Specifically, given a user's personalization context and a task description for the personalized LLM, we examine the model's attention patterns across its context. Suppose that we have a long user history/context $H$ (e.g., the user's movie interaction history or writing history) and a task description $\mathcal{T}$ (e.g., recommend next movie or suggest title). We input these to a marking LLM $\Phi_{\text{Mark}}$. Following previous work \cite{}, we focus on the attention scores of the last input tokens for the user context, as these directly contribute to the model's answer. 

\paragraph{Token-level attention scores.}
Suppose that there are $N$ input tokens in total. We calculate the attention scores of the last token to the input sequence at the layer $d$. Denote the token-level attention probability vector across input tokens of layer $d$ and head $j$  as $\mathbf{w}^{(d,j)} := [w_1^{(d,j)}, w_2^{(d,j)}, \dots, w_N^{(d,j)}] \in \mathbb{R}^N$. The average attention scores across all heads can be defined as:
\begin{equation}
\label{eq:token_attn}
\tilde{\mathbf{w}}^{(d)} = \frac{1}{J} \sum_{j=1}^{J} \mathbf{w}^{(d,j)}
\end{equation}
where $J$ is the number of attention heads and $\mathbf{w}^{(d,j)} \in \mathbb{R}^N$ is the attention vector for head $j$. Thus, $\tilde{\mathbf{w}}^{(d)} := [\tilde{w}_1^{(d)}, \tilde{w}_2^{(d)}, \dots, \tilde{w}_N^{(d)}] \in \mathbb{R}^N$. 

\paragraph{Sentence-level attention scores.}
After obtaining the token-level attention scores $\tilde{\mathbf{w}}^{(d)}$, we aggregate and average these scores across sentences to derive sentence-level attention scores. For sentence $u_i$ spanning tokens from position $p_{start}^{u_i}$ to $p_{end}^{u_i}$, the sentence-level attention score is:
\begin{equation}
\label{eq:sentence_attn}
 \hat{w}_i^{(d)} = \frac{1}{p_{end}^{u_i} - p_{start}^{u_i} + 1} \sum_{m=p_{start}^{u_i}}^{p_{end}^{u_i}} \tilde{w}_m^{(d)}
\end{equation}

\paragraph{Signal-level attention scores.}
To systematically analyze which personalization signals matter most, we categorize user behavioral data into distinct signal types $\mathcal{L}_s = \{\tau_1, \tau_2, \ldots, \tau_K\}$ (e.g., titles, user ratings, user reviews), where $K$ is the number of signal types. Each sentence $u_i$ in the user context is assigned to exactly one signal type $\tau_k \in \mathcal{S}$. To reveal which signal types receive the most attention, we compute type-wise average scores:
\begin{equation}
\label{eq:signal_attn}
\hat{w}_{\tau_k}^{(d)} = \frac{1}{|\mathcal{U}_{\tau_k}|} \sum_{u_i \in \mathcal{U}_{\tau_k}} \hat{w}_i^{(d)}
\end{equation}
where $\mathcal{U}_{\tau_k}$ represents all sentences of type $\tau_k$.

\subsection{Key Findings}
\label{Sec:Findings}

\begin{figure*}[t]
    \centering
    \includegraphics[width=1\textwidth]{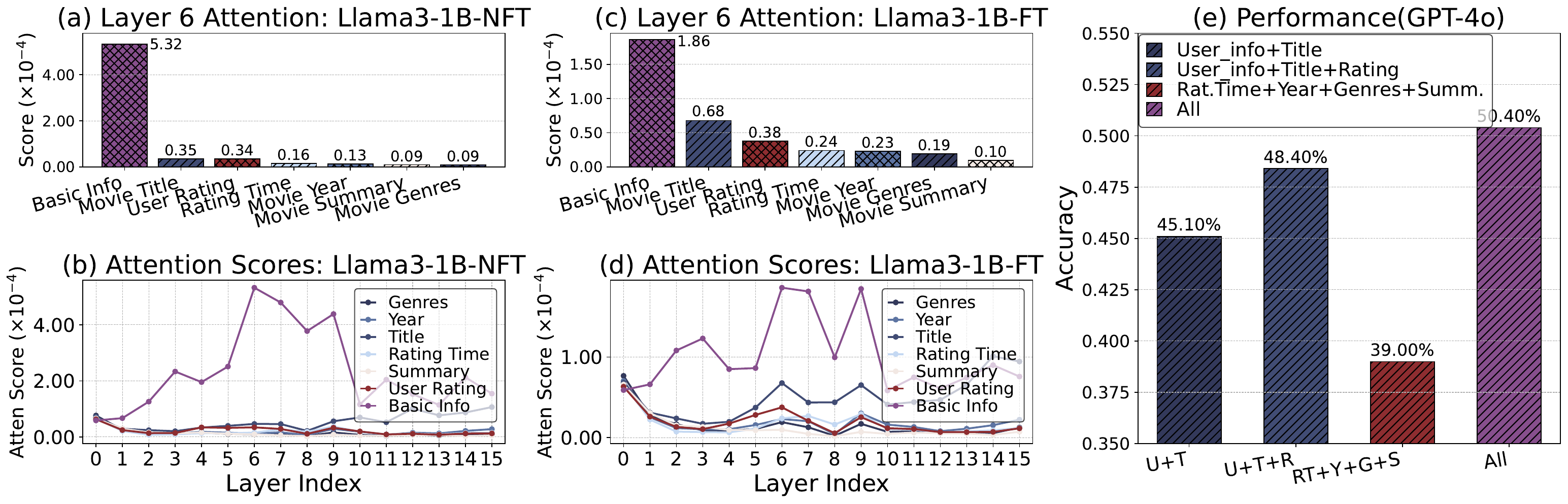}
    
    \caption{Attention visualization on MovieLens dataset. Layer-6 attention and cross-layer attention for non-fine-tuned $\Phi_{\text{Mark}}$ (a-b) and fine-tuned $\Phi_{\text{Mark}}$ (c-d), and performance comparison across signal subsets (e). U: User Basic Info, T: Title, R: User Rating, RT: Rating Time, Y: Movie Year, G: Genre, S: Movie Summary, All: All signals.}
    \label{fig:pre_movie}
\end{figure*}

\begin{figure*}[t]
    \centering
    \includegraphics[width=1\textwidth]{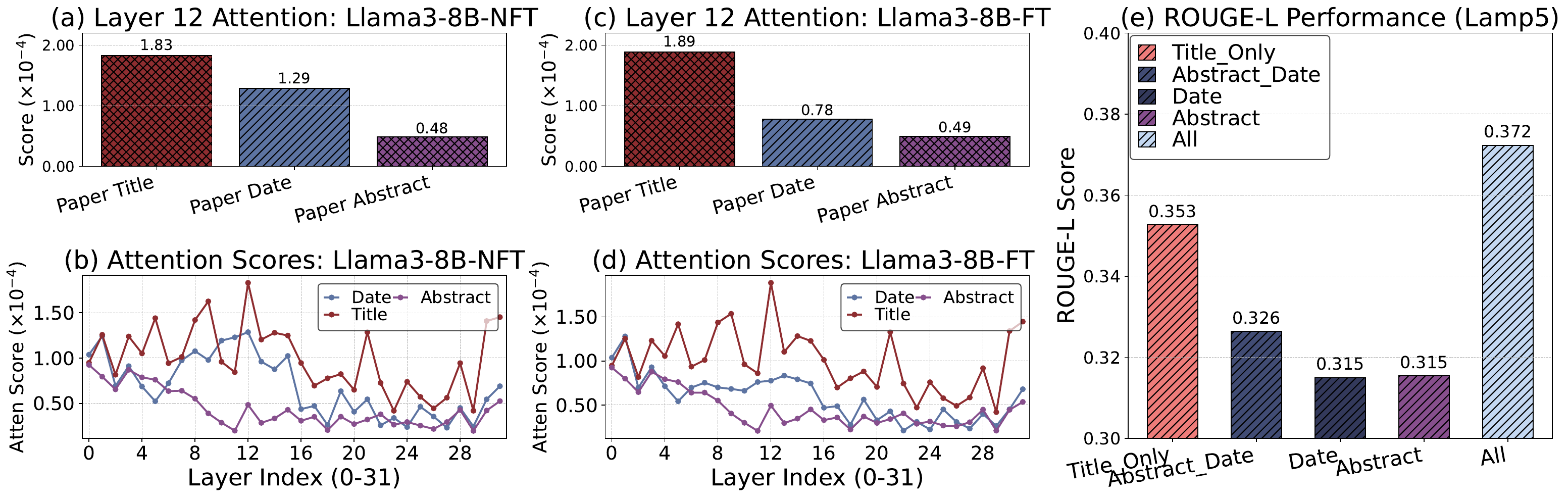}
    \caption{Attention visualization on LaMP-5 dataset. Layer-12 attention and cross-layer attention for non-fine-tuned $\Phi_{\text{Mark}}$ (a-b) and fine-tuned $\Phi_{\text{Mark}}$ (c-d), and performance comparison across signal subsets (e).}
    \label{fig:pre_lamp}
    \vspace{-0.5cm}
\end{figure*}

To explore whether LLMs' attention scores reflect personalization signal importance, we conduct a study on 2 representative personalization datasets. Specifically, we examine whether the model assigns higher attention to important signal types and lower attention to unimportant ones.

\paragraph{Datasets.} We conduct this study on two datasets:  \textbf{(a) MovieLens-1M}~\cite{10.1145/2827872} for movie recommendation. Signal types withine contexts include movie titles, user ratings, summaries, genres, watch times, release years, and users' basic information(age, gender, occupation). Following prior content-based recommendation work~\cite{zhang2024agentcf,zhang2024generative}, the task is to select the user's last interaction from among four random negatives, with all previous interactions serving as personalized context. We use 1,000 users for training the Mark model and 1,000 users for testing and attention visualization.  \textbf{(b) LaMP-5}~\cite{salemi2024lamp} for personalized title generation. Contexts contain authors' previous papers, including titles, abstracts, and publication dates. The task is to generate personalized paper titles given an abstract and the author's writing history. We use 750 users for training the Mark model and 750 users for testing and attention visualization. Detailed dataset descriptions and examples are provided in  Appendix~\ref{sec:dataset}.

\paragraph{Experimental Setup.} As introduced in Section~\ref{Sec:Description}, we input the user personalized context $H$ and task description $\mathcal{T}$ of each dataset into the mark model $\Phi_{\text{Mark}}$ to obtain attention weights $\hat{w}_{\tau_k}^{(d)}$ for different signal types. For MovieLens, we use both the original Llama-3.1-1B-Instruct and its fine-tuned version as mark models. For LaMP-5, we employ both the original Llama-3.1-8B-Instruct and its fine-tuned version as mark models. Note that the fine-tuned $\Phi_{\text{Mark}}$ is obtained by fine-tuning the original model on the training set following standard practices, where the input consists of the concatenation of $H$, $\mathcal{T}$, and the actual question (i.e., candidate movies for MovieLens or given abstract for LaMP-5), with the ground-truth answer as the output target. To validate the correlation between attention scores and signal importance, we also evaluate performance using only subsets of signals (important signals only vs. unimportant signals only) with GPT-4o as the generator. Performance is measured by accuracy for MovieLens-1M and ROUGE-L for LaMP-5.

\paragraph{Results \& findings.}

The results for the MovieLens dataset are shown in Figure~\ref{fig:pre_movie}.  Figures~\ref{fig:pre_movie}(b) and~\ref{fig:pre_movie}(d) show signal-level attention distributions across layers for different signal types, while Figures~\ref{fig:pre_movie}(a) and~\ref{fig:pre_movie}(c) display the attention distributions at layer 6. Attention scores begin to diverge in the first few layers and exhibit relatively large differences in the early-to-middle layers (e.g., layer 6).  For the original model (Figures~\ref{fig:pre_movie}(a) and~\ref{fig:pre_movie}(b)), the model primarily attends to users' basic information while assigning low attention to other signal types, with slightly higher attention to movie titles and user ratings. In contrast, the finetuned model(Figures~\ref{fig:pre_movie}(c) and~\ref{fig:pre_movie}(d)) demonstrates clearer discriminative attention across signal types. While user basic information still receives the highest attention (though reduced compared to the original model), among the remaining signals, movie titles receive substantially higher attention, followed by user ratings, with other signals receiving notably less attention. This attention pattern aligns closely with the performance results in Figure~\ref{fig:pre_movie}(e), where using only the important signals identified by attention scores achieves significantly higher accuracy (Basic+Title [U+T]: 45.1\%, Basic+Title+Rating [U+T+R]: 48.4\%) compared to using all low-attention signals (Rating Time+Year+Genre+Summary [RT+Y+G+S]: 39.0\%).


 We observe similar patterns on the LaMP-5 dataset in Figure~\ref{fig:pre_lamp}. The attention scores show large differences in the early-to-middle layers (e.g., layer 12). The original model assigns high attention to paper titles and publication dates (with minimal distinction between them) and low attention to abstracts. For the finetuned model, title signals (important signals) become dominantly higher than the other two signal types (unimportant signals). This pattern better aligns with the performance results in Figure~\ref{fig:pre_lamp}(e), where using only title information yields clearly higher performance than using paper dates, abstracts, or both together.

These results preliminarily demonstrate that \textit{(a) LLMs' attention scores can naturally reveal the importance of personalization signals} and that \textit{(b) finetuning enhances the model's ability to distinguish between important and unimportant signals}.\footnote{This claim is further validated in Section~\ref{Sec:Ablation}.} This finding motivates us to utilize attention scores to mark important signals for better summarization.

\section{Method}
\label{Sec:Method}

\begin{figure}[h]
    \centering
    \includegraphics[width=1\linewidth]{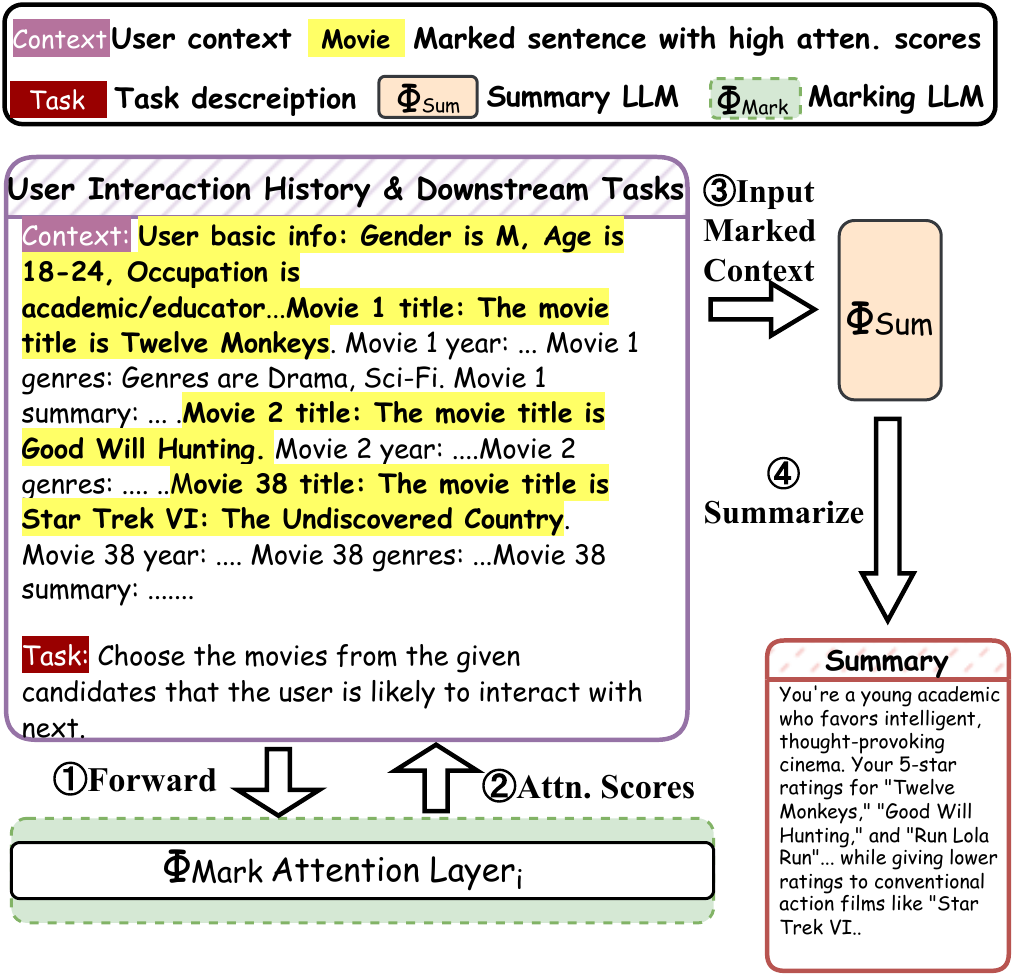}
    
    \caption{An illustration of Attn-GS framework.   }
    \vspace{-0.6cm}
    \label{fig:method}
    
\end{figure}

Inspired by the above findings, we propose an attention-guided summarization pipeline \textit{Attn-GS} to utilize LLMs' attention feedback for generating high-quality compressed user contexts. The overall framework is illustrated in Figure~\ref{fig:method} and consists of two stages: (1) \textbf{Critical Personalization Sentence Marking}, which leverages attention feedback from a white-box marking model $\Phi_{\text{Mark}}$ to identify and highlight important personalization sentences based on attention scores, and (2) \textbf{Summarization Based on Marked Context}, which utilizes a summary model $\Phi_{\text{Sum}}$ that takes the marked context as input to generate a compressed profile within a predefined token limit. We introduce these two stages in detail in Sections~\ref{Sec:Method_Marking} and \ref{Sec:Method_Summarize}, respectively, and present the full pipeline in Section~\ref{Sec:method_Alg}\footnote{Examples of Marked Sentence and summarized results can be found in Appendix~\ref{Sec:Marked contexts} and \ref{Sec:Summarization Resukts}}.





\subsection{Critical Personalization Sentence Marking}
\label{Sec:Method_Marking}
To identify which personalization sentences contain critical information, we first input the user's history context concatenated with the task description into the marking model $\Phi_{\text{Mark}}$ to obtain token-level attention scores $\tilde{\mathbf{w}}^{(d)}$, as shown in Eq.(\ref{eq:token_attn}):
\begin{equation}
\label{eq:attn_process}
\tilde{\mathbf{w}}^{(d)} \leftarrow \text{AttnScores}(\Phi_{\text{Mark}}(H, \mathcal{T}), d)
\end{equation}

Next, we aggregate the token-level attention scores into sentence-level attention scores $\hat{w}_i^{(d)}$ for every sentence $u_i \in H$. We then select sentences with relatively high attention scores using a threshold-based filtering mechanism. Specifically, given a threshold factor $\alpha \in (0,1]$, we identify the set of important sentences as:
\begin{equation}
\label{eq:important_sentence}
\mathcal{S}_{\text{Mark}} = \left\{u_i \in H \mid \hat{w}_i^{(d)} \geq \alpha \cdot \max_{u_j \in H} \hat{w}_j^{(d)}\right\}
\end{equation}
where sentences whose attention scores exceed $\alpha$ times the maximum attention score are marked as important.

After identifying these important personalization sentences, we explicitly mark them to obtain a modified version of the user context, denoted as $H^*$.  Following previous practices ~\cite{liu2025selfelicit}, we add markers \texttt{<start\_important>} before each important sentence and \texttt{<end\_important>} after it. The entire marking process can be expressed as:
\begin{equation}
\label{eq:Mark}
H^* \leftarrow \text{Mark}(\mathcal{S}_{\text{Mark}}, H)
\end{equation}

This marking helps the subsequent summary model become aware of and prioritize these personalization signals during summarization.

\subsection{Summarization Based on Marked Context}
\label{Sec:Method_Summarize}
After obtaining the marked version of the user context $H^*$ with highlighted personalization sentences $\mathcal{S}_{\text{Mark}}$, we generate a compressed user profile $G$ within a token limit of $m$ tokens. We input $H^*$ into the summary model $\Phi_{\text{Sum}}$ with a prompt that explicitly instructs the model to prioritize these important sentences when generating the compressed profile. Example prompt templates can be seen in Appendix~\ref{Sec:Prompts}. The overall summarization process can be expressed as: G $\leftarrow \Phi_{\text{Sum}}(H^*, m)$.

\subsection{Attn-GS Algorithm}
\begin{algorithm}
\caption{Attn-GS Algorithm}
\label{alg:upm}
\textbf{Input:} Language Models $\Phi_{\text{Mark}}$ and $\Phi_{\text{Sum}}$, User History $H$, Task Description $\mathcal{T}$, Attention Layers $\mathcal{L}_{\text{Attn}}$, Target Layer Index $d$, Threshold $\alpha$, Token limits $m$.
\textbf{Output:} User Profile Summary $G$.

\begin{algorithmic}[1]
\State // \textit{Step 1: Compute Attention Scores}
\State $\mathbf{w}^{(d)} \leftarrow \text{AttnScores}(\Phi_{\text{Mark}}(H, \mathcal{T}), d)$ \Comment{Eq.(\ref{eq:attn_process})}
\State // \textit{Step 2: Select Key Information}
\State $S_{\text{Mark}} \leftarrow \text{SelectSentences}(\mathbf{w}^{(d)}, H, \alpha)$ \Comment{Select  $S_{\text{Mark}}$ with high attn scores (Eq.(\ref{eq:important_sentence}))}
\State // \textit{Step 3: Construct Input Context}
\State $H^* \leftarrow \text{Mark}(S_{\text{Mark}}, H)$ \Comment{highlighting $S_{\text{Mark}}$}
\State // \textit{Step 4: Generate Summary with Sum LLM}
\State $G \leftarrow \Phi_{\text{Sum}}(H^*,m)$ 
\State \textbf{return} $G$
\end{algorithmic}
\end{algorithm}

\label{Sec:method_Alg}
The framework's process can be summarized  in Algorithm~\ref{alg:upm}. First, we input the user interaction history and task description to the marking model $\Phi_{\text{Mark}}$  to obtain attention scores(step 1)  and identify(step 2) and mark(step 3) critical personalization sentences with high attention scores. Next, we input the marked context to the summary model to generate high-quality, task-relevant compressed context(step 4). This entire process can be conducted offline. During inference, the compressed context can be combined with user real-time queries(e.g, movie candidates) as input to generator $\Phi_{\text{G}}$ to reduce inference latency and costs.

\section{Experiment}
\label{Sec:experiment}
In this section, we conduct comprehensive experiments to validate the effectiveness of our proposed Attn-GS method. We first introduce our experimental settings in Section~\ref{Sec:Ex_setting}, then compare the performance of our method against traditional baselines in two common personalization settings: (a) \textbf{Inference-only}: using the compressed profile for direct inference (Section~\ref{Sec:Ex_Zeroshot}), and (b) \textbf{Training and inference}: using the compressed profile from a subset of users for model training and inference on another subset (Section~\ref{Sec:Ex_FT}). Finally, we conduct further probing on token efficiency, fine-tuned vs. non-fine-tuned mark models, threshold $\alpha$, and mark model layer selection in Section~\ref{Sec:Ablation}.


\begin{table*}[h!]
\centering
\resizebox{0.9\textwidth}{!}{
\begin{tabular}{lcccccccc}
\toprule
\multirow{2}{*}{\textbf{Method}} & \multicolumn{8}{c}{\textbf{Token Count}} \\
\cmidrule(lr){2-9}
& \multicolumn{2}{c}{\textbf{50}} & \multicolumn{2}{c}{\textbf{100}} & \multicolumn{2}{c}{\textbf{150}} & \multicolumn{2}{c}{\textbf{200}} \\
\cmidrule(lr){2-3} \cmidrule(lr){4-5} \cmidrule(lr){6-7} \cmidrule(lr){8-9}
& \textbf{Llama-3} & \textbf{GPT-4o-mini} & \textbf{Llama-3} & \textbf{GPT-4o-mini} & \textbf{Llama-3} & \textbf{GPT-4o-mini} & \textbf{Llama-3} & \textbf{GPT-4o-mini} \\
\midrule
Truncate & 40.9 & 41.0 & 41.9 & 42.6 & 42.3 & 43.2 & 42.5 & 45.8 \\
Direct Summary & 41.2 & 43.2 & 43.0 & 45.4 & 43.5 & 45.6 & 44.1 & 45.7 \\
Random Mark & 41.5 & 42.8 & 42.6 & 44.9 & 43.1 & 45.2 & 43.7 & 45.3 \\
Mark All & 41.8 & 43.5 & 43.2 & 45.6 & 43.7 & 45.8 & 44.3 & 45.9 \\
Prompt-G & 41.6 & 43.4 & 43.1 & 45.5 & 43.8 & 46.0 & 44.4 & 46.1 \\
CoT & 41.4 & 43.1 & 42.8 & 45.2 & 43.6 & 45.8 & 44.2 & 46.0 \\
Self-Reflection & 41.3 & 43.3 & 43.0 & 45.3 & 43.4 & 45.7 & 44.0 & 45.8 \\
\textbf{Attn-G} & \textbf{44.7} & \textbf{45.3} & \textbf{46.0} & \textbf{48.0} & \textbf{47.1} & \textbf{49.6} & \textbf{47.7} & \textbf{50.6} \\
\bottomrule
\end{tabular}%
}
\caption{Zero-Shot Accuracy (\%) on MovieLens. Full context: Llama-3 48.7\%, GPT-4o-mini 52.4\%. No context: Llama-3 19.8\%, GPT-4o-mini 21.1\%.}

\label{tab:zero-shot-performance-movielens}
\end{table*}

\begin{table*}[h!]
\centering
\resizebox{0.9\textwidth}{!}{%
\begin{tabular}{lcccccccc}
\toprule
\multirow{2}{*}{\textbf{Method}} & \multicolumn{8}{c}{\textbf{Token Count}} \\ 
\cmidrule(lr){2-9} 
& \multicolumn{2}{c}{\textbf{50}} & \multicolumn{2}{c}{\textbf{100}} & \multicolumn{2}{c}{\textbf{150}} & \multicolumn{2}{c}{\textbf{200}} \\ 
\cmidrule(lr){2-3} \cmidrule(lr){4-5} \cmidrule(lr){6-7} \cmidrule(lr){8-9} 
& \textbf{Llama-3} & \textbf{GPT-4o-mini} & \textbf{Llama-3} & \textbf{GPT-4o-mini} & \textbf{Llama-3} & \textbf{GPT-4o-mini} & \textbf{Llama-3} & \textbf{GPT-4o-mini} \\ 
\midrule 
Truncated Cont & 0.330 & 0.353 & 0.343 & 0.368 & 0.344 & 0.370 & 0.345 & 0.370 \\
Direct Summary & 0.338 & 0.371 & 0.352 & 0.375 & 0.353 & 0.377 & 0.354 & 0.378 \\
Random Mark & 0.334 & 0.367 & 0.348 & 0.371 & 0.349 & 0.373 & 0.350 & 0.375 \\
Mark All & 0.339 & 0.372 & 0.352 & 0.376 & 0.352 & 0.376 & 0.354 & 0.378 \\
Prompt-G & 0.338 & 0.371 & 0.352 & 0.376 & 0.354 & 0.378 & 0.355 & 0.379 \\
CoT & 0.336 & 0.370 & 0.349 & 0.374 & 0.351 & 0.377 & 0.352 & 0.379 \\
Self-Reflection & 0.335 & 0.370 & 0.349 & 0.375 & 0.350 & 0.376 & 0.351 & 0.379 \\
\textbf{Attn-G} & \textbf{0.359} & \textbf{0.388} & \textbf{0.370} & \textbf{0.394} & \textbf{0.372} & \textbf{0.397} & \textbf{0.374} & \textbf{0.401} \\
\bottomrule
\end{tabular}%
}
\caption{ROUGE-L/Lsu Results. \textbf{Baselines:} The upper bound ('Full Context') for Llama-3-8B is 0.3895 and for GPT-4o-mini is 0.4139. The lower bound ('None') for Llama-3-8B is 0.256 and for GPT-4o-mini is 0.316.}
\label{tab:Zero-shot_Lamp-performance}
\vspace{-0.5cm}
\end{table*}

\subsection{Experiment Settings}
\paragraph{Datasets.} 
We evaluate our method on two datasets: MovieLens-1M for context-based movie recommendation and LaMP-5 for personalized title generation. Following the settings in Section~\ref{Sec:Findings}, the MovieLens task selects the user's truly interacted movie from 5 candidates (measured by accuracy), while the LaMP-5 task generates personalized paper titles from abstracts (measured by ROUGE-L scores).  For inference-only settings, we evaluate on 1,000 MovieLens users and 750 LaMP-5 users using compressed user contexts. For training and inference settings, we train the generation model $\Phi_{\text{G}}$ on 500 MovieLens users and 375 LaMP-5 users with compressed contexts, then test on another 500 and 350 users, respectively. Note that all experimental data is excluded from the marking model $\Phi_{\text{Mark}}$ training process.

\paragraph{Models.} 

We use the fine-tuned models from Section~\ref{Sec:Findings} as $\Phi_{\text{Mark}}$ for both datasets. We use Layer 6 (Llama-3.1-1B) for MovieLens marking and Layer 12 (Llama-3.1-8B) for LaMP marking, with the threshold set to 0.2 by default.  For inference-only settings, we explore two configurations: (1) Llama-3.1-8B-Instruct as both $\Phi_{\text{Sum}}$ and $\Phi_{\text{G}}$, and (2) GPT-4o-mini as both $\Phi_{\text{Sum}}$ and $\Phi_{\text{G}}$. For training and inference settings, we use Llama-3.1-8B-Instruct as $\Phi_{\text{Sum}}$ and Llama-3.1-1B-Instruct as $\Phi_{\text{G}}$.


\paragraph{Baselines.} To validate the effectiveness of our method and the necessity of attention-based marking, we compare against various baselines across four categories:  \textbf{Truncation-based:} (a) \textit{Truncate}~\cite{everaert2024amazonqac} directly uses the most recent user contexts. \textbf{Direct summarization:} (b) \textit{Direct Summary}~\cite{zhang2024generative} inputs unmarked contexts into $\Phi_{\text{Sum}}$ to derive compressed profiles. \textbf{Reasoning-enhanced:} (c) \textit{CoT}~\cite{wei2022chain} prompts $\Phi_{\text{Sum}}$ to think step-by-step for summarization; (d) \textit{Self-Reflection}~\cite{ji2023towards} asks $\Phi_{\text{Sum}}$ to self-reflect and refine its summary. \textbf{Alternative marking methods:} (e) \textit{Mark All} marks all sentences as important for summarization; (f) \textit{Random Mark} randomly marks the same number of sentences as $H^*$ for summarization; (g) \textit{Prompt-GS} first prompts $\Phi_{\text{Sum}}$ to identify important sentences, marks them to derive $H^*$, then summarizes.\footnote{More detail of these baselines are shown in Appendix~\ref{Sec:baselines}.} We compare our Attn-GS method with these baselines under various maximum token settings [50, 100, 150, 200]. The prompt template for inference and training for  $\Phi_{\text{G}}$ is fixed for fair comparison. 


\label{Sec:Ex_setting}

\subsection{Inference-only Performance}
In this subsection, we evaluate the performance of Attn-GS in the inference-only setting, where compressed user contexts are directly used with real-time queries for inference. Results on the MovieLens and LaMP-5 datasets are shown in Tables~\ref{tab:zero-shot-performance-movielens} and~\ref{tab:Zero-shot_Lamp-performance}, respectively.  From Table~\ref{tab:zero-shot-performance-movielens}, we observe that utilizing personalized contexts is essential for task performance, and performance generally increases with token length. Among all baselines, Truncate and Direct Summary consistently yield unsatisfactory performance, while reasoning-enhanced methods (CoT and Self-Reflection) do not provide clear performance gains, demonstrating that simply relying on LLMs' reasoning ability is insufficient to generate effective compressed profiles. Alternative marking methods such as Mark All, Random Mark, and Prompt-GS do not effectively enhance performance compared with Direct Summary, indicating their inability to identify truly important sentences without utilizing attention signals.  In contrast, our Attn-GS consistently achieves the best performance across all settings (models and token lengths) and scales well with increasing tokens. Attn-GS achieves performance within 1.8\% of using full context (10,000 tokens) with only 200 tokens, representing a 50$\times$ reduction. Similar findings are observed in Table~\ref{tab:Zero-shot_Lamp-performance}, where Attn-GS significantly outperforms all baselines and achieves performance close to using full context. These findings validate the necessity of attention-based marking and the effectiveness of our method in inference-only scenarios.

\begin{table}[htpb]
\centering
\resizebox{0.7\linewidth}{!}{%
\begin{tabular}{lcccc} 
\toprule
\multirow{2}{*}{\textbf{Method}} & \multicolumn{4}{c}{\textbf{Token Count}} \\ 
\cmidrule(lr){2-5}
& \textbf{50} & \textbf{100} & \textbf{150} & \textbf{200} \\ 
\midrule
Truncate & 52.60 & 54.00 & 56.00 & 58.10 \\
Direct Summary & 56.20 & 57.10 & 58.40 & 59.80 \\
Random Mark & 55.80 & 56.70 & 58.00 & 59.40 \\
Mark ALL & 56.40 & 57.30 & 58.60 & 59.90 \\
Prompt-G & 56.00 & 56.90 & 58.20 & 59.60 \\
CoT & 55.90 & 56.80 & 58.10 & 59.50 \\
Self-Reflection & 56.10 & 57.00 & 58.30 & 59.60 \\
\textbf{Attn-G} & \textbf{58.20} & \textbf{59.00} & \textbf{61.80} & \textbf{64.60} \\
\bottomrule
\end{tabular}%
}
\caption{FT performance on MovieLens (accuracy \%): Llama-1B-FT. Full context: 67.20\%.}
\vspace{-0.3cm}
\label{tab:ft-performance-llama-v4}
\end{table}

\label{Sec:Ex_Zeroshot}
\subsection{Training and Inference Setting Performance}
\label{Sec:Ex_FT}

In this subsection, we investigate the performance of Attn-GS in the training and inference setting. In this setting, compressed user profiles are utilized to train a better $\Phi_{\text{G}}$ that can be used for further inference (also taking compressed profiles as input). The results are presented in Table~\ref{tab:ft-performance-llama-v4} for MovieLens and Table~\ref{tab:lamp-performance-final-v7} for LaMP-5.  From these tables, we observe that while increasing token count enhances performance, all baselines (truncation-based, direct summarization, reasoning-enhanced, and alternative marking) yield unsatisfactory performance and present a large performance gap compared to using full context. In contrast, our Attn-GS significantly outperforms all baselines across all token settings, with a much smaller gap to using full context. These results demonstrate the superiority of Attn-GS not only in inference-only settings but also for model training, and further validate that the compressed user contexts generated by Attn-GS contain more valuable key personalization information for downstream tasks.

\begin{table}[htpb]
\centering
\resizebox{0.7\linewidth}{!}{
\begin{tabular}{lcccc} 
\toprule
\multirow{2}{*}{\textbf{Method}} & \multicolumn{4}{c}{\textbf{Token Count}} \\ 
\cmidrule(lr){2-5} 
& \textbf{50} & \textbf{100} & \textbf{150} & \textbf{200} \\ 
\midrule 
Truncated Cont & 0.382 & 0.384 & 0.385 & 0.386 \\
Simple Summary & 0.384 & 0.384 & 0.386 & 0.389 \\
Random Mark & 0.380 & 0.381 & 0.384 & 0.387 \\
Mark ALL & 0.385 & 0.384 & 0.386 & 0.389 \\
Prompt & 0.384 & 0.384 & 0.386 & 0.390 \\
CoT & 0.383 & 0.383 & 0.386 & 0.388 \\
Self-Reflection & 0.383 & 0.383 & 0.385 & 0.389 \\
\textbf{Attn-GS} & \textbf{0.396} & \textbf{0.402} & \textbf{0.404} & \textbf{0.405} \\
\bottomrule
\end{tabular}%
}
\caption{FT Performance on LaMP-5(Rouge-L) 'Full Context': 0.4147.}
\label{tab:lamp-performance-final-v7}
\vspace{-0.5cm}
\end{table}

\begin{figure}[t]
\centering
\resizebox{\linewidth}{!}{%
    \begin{minipage}{\linewidth}
        \subfloat[Ablation study on $\alpha$]{\includegraphics[width=.5\linewidth]{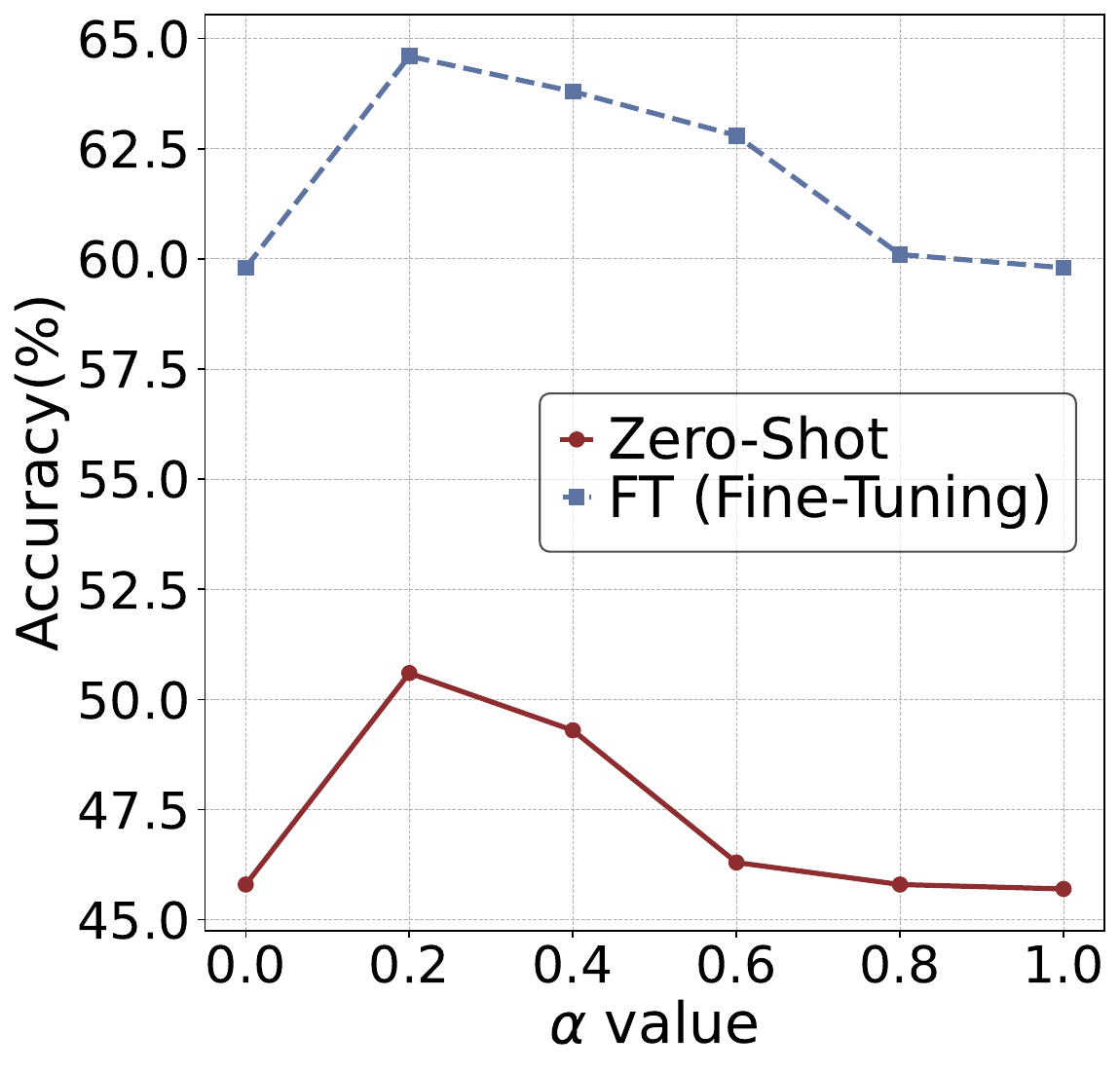}
        \label{fig:alpha}}
        \subfloat[Ablation study on layer]{\includegraphics[width=.49\linewidth]{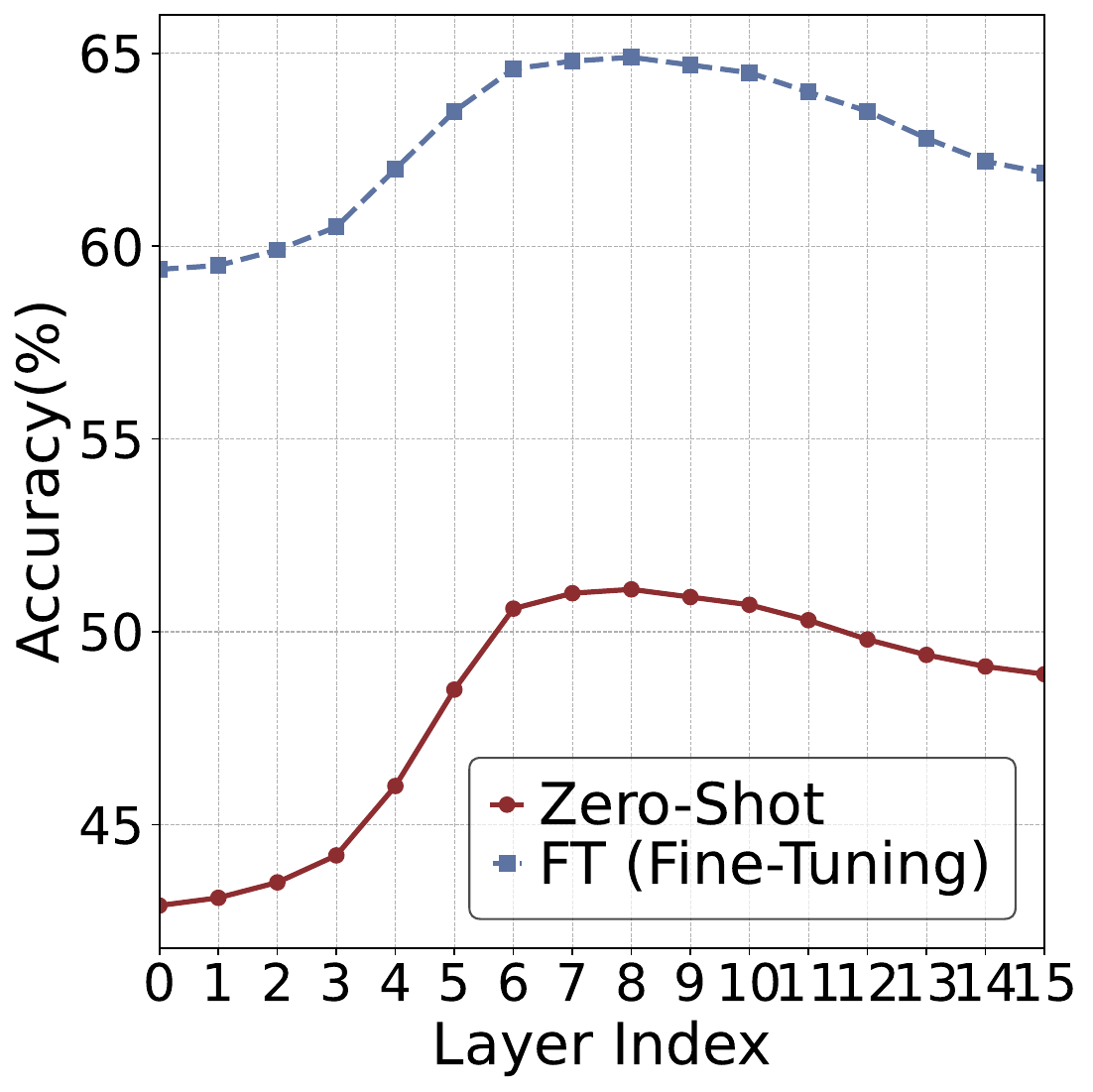}
        \label{fig:Layers}}
    \end{minipage}
}
\caption{Ablation Studies}
\vspace{-0.3cm}
\label{fig:Ablation Study}
\end{figure}
\subsection{Further Probing}
\label{Sec:Ablation}

\begin{table}[h!]
\centering

\resizebox{1\linewidth}{!}{
\begin{tabular}{lcccc}
\toprule
\textbf{Method} & \textbf{Truncate} & \textbf{Direct Summ.} & \textbf{Prompt-GS} & \textbf{Attn-GS} \\
\midrule
Tokens & 750 & 750 & 700 & \textbf{100} \\
Percentage  & 8.0\% & 8.0\% & 7.5\% & \textbf{1.1\%} \\
\bottomrule
\end{tabular}
}
\caption{Efficiency Comparison(GPT-4o-mini,48\%)}
\vspace{-0.5cm}
\label{tab:token-efficiency}
\end{table}

\begin{table}[htpb]
\centering
\resizebox{0.8\linewidth}{!}{
\begin{tabular}{lcccc}
\toprule
\textbf{Method} & \textbf{50} & \textbf{100} & \textbf{150} & \textbf{200} \\
\midrule
Not Fine-tuned $\Phi_{\text{Mark}}$ & 44.1 & 46.2 & 47.1 & 47.9 \\
Fine-tuned $\Phi_{\text{Mark}}$ & 45.3 & 48.0 & 49.6 & 50.6 \\
\bottomrule
\end{tabular}
}
\vspace{-0.2cm}
\caption{Fine-tuned vs. Not Fine-tuned $\Phi_{\text{Mark}}$}
\label{tab:finetuned-comparison}
\vspace{-0.7cm}
\end{table}

\paragraph{Token Efficiency.} To quantify Attn-GS's token efficiency, we compare the number of tokens required to achieve 48\% accuracy in the inference-only setting with GPT-4o-mini as $\Phi_{\text{G}}$ in Table~\ref{tab:token-efficiency}. We observe that Attn-GS reduces token requirements by 7$\times$ compared to  baselines, using only 1.1\% of full context tokens, which validates that Attn-GS significantly reduces inference tokens.
\vspace{-0.2cm}
\paragraph{Finetuned mark model vs not finetuned.} To validate that a fine-tuned $\Phi_{\text{Mark}}$ is more suitable for context marking than a non-fine-tuned $\Phi_{\text{Mark}}$, we compare their performance in Table~\ref{tab:finetuned-comparison} in the inference-only setting with GPT-4o-mini as $\Phi_{\text{G}}$ and other parameters fixed. We observe that although the non-fine-tuned model still outperforms baselines in Table~\ref{tab:zero-shot-performance-movielens}, it clearly underperforms Attn-GS using the fine-tuned model as $\Phi_{\text{Mark}}$. This further validates the findings in Section~\ref{Sec:Findings} that fine-tuning enhances the model's ability to distinguish between important and unimportant signals.
\vspace{-0.3cm}
\paragraph{Thereshold $\alpha$.}
We fix other parameters and vary the marking threshold $\alpha$ for MovieLens compression, reporting the downstream inference-only performance at 200 tokens on GPT-4o-mini as well as training and inference performance on Llama-1B-Instruct  in Figure~\ref{fig:alpha}. From the results, we observe that choosing a threshold between 0.2 and 0.4 yields the best overall performance. Setting the threshold too high (e.g., marking only the most highly-attended sentences) or too low (e.g., Mark All) fails to effectively guide the summarization process. We recommend carefully selecting a moderate threshold value based on validation set performance in practice.
\vspace{-0.3cm}
\paragraph{Marking layers.} Similarly, we keep other parameters fixed and vary the model layer of $\Phi_{\text{Mark}}$, reporting both downstream inference-only performance at 200 tokens on GPT-4o-mini (as $\Phi_{\text{Sum}}$ and $\Phi_{\text{G}}$) and training and inference performance (Llama-3.1-8B-Instruct as $\Phi_{\text{Sum}}$ and Llama-3.1-1B-Instruct as $\Phi_{\text{G}}$). From the results in Figure~\ref{fig:Layers}, we observe that early layers yield low performance, performance peaks in the middle layers, and slightly decreases in the final layers. This is likely because early layers primarily capture low-level syntactic features, while middle and later layers are more responsible for processing personalized information and allocating attention weights for generation. Based on the results, we recommend using middle layers of LLMs for marking.

\vspace{-0.2cm}
\section{Conclusions}
\label{Conclusion}
\vspace{-0.2cm}

In this work, we explore the potential of leveraging LLM attention patterns to guide personalization context compression. Through extensive experiments, we discover that LLMs' attention patterns naturally reveal important personalization signals, and fine-tuning enhances their ability to distinguish relevant information. Based on these findings, we propose Attn-GS, an attention-guided compression framework that uses fine-tuned attention patterns to intelligently compress personalization contexts. Experiments demonstrate that Attn-GS consistently outperforms all baselines, achieving close performance of using full context while reducing token usage by 50$\times$ compared to full context and 7$\times$ compared to baseline methods. 

\section*{Limitations}



While our work demonstrates the potential of attention-guided compression for personalization tasks, several aspects warrant further exploration. First, although we reveal that attention patterns can unveil personalization signal importance, a deeper understanding of how LLMs internally process and integrate personalization inputs remains an open question. Future work could investigate the underlying mechanisms in greater detail to develop more sophisticated compression strategies. Second, extending our framework to broader scenarios, including multi-task settings and diverse application domains, would help establish the generalizability and scope of attention-guided compression approaches.

\bibliography{anthology}

\clearpage

\appendix

\section{Appendix}
\subsection{Dataset Details}
\label{sec:dataset}
\paragraph{MovieLens-1M.}In our experiment, we utilize a subset of MovieLens-1M dataset~\cite{10.1145/2827872}, using 1000 users for training while another 1000 users for testing. For each user, user contexts include Movie Title, User Rating, Movie Summary, Movie Genres, Watch Time, Movie Year, and User Basic Information (age, gender, occupation), providing diverse personalization signals for analysis. Examples of user contexts can be found at Figure~\ref{fig:movielens_context}.

\paragraph{LaMP-5.} In our experiment we utilize a subset of LaMP-5: Personalized Scholarly Title
Generation dataset~\cite{salemi2024lamp} . The user contexts include the authors' previous papers, with signals comprising paper title, paper summary, and publication date. Examples can be found in Figure~\ref{fig:lamp5_context}. For each user, the task is to generate a personalized paper title based on a given paper summary and the user's writing history. We measure performance using Rouge-L scores between the generated titles and ground truth (actual titles).

\subsection{Details of Attn-GS}
\label{Sec:attention guided summarization}
\subsubsection{Prompt Used}
\label{Sec:Prompts}
The summarization prompts used for MovieLens are shown in Figure~\ref{fig:movie_prompt_template} , used for Lamp-5 are shown in Figure~\ref{fig:academic_title_prompt}.
\subsubsection{Examples of Marked Contexts}
\label{Sec:Marked contexts}
Examples of Marked Contexts are shown in Figures~\ref{fig:movielens_context} and~\ref{fig:lamp5_context}. The results demonstrate that LLMs' attention mechanisms can effectively identify important personalization signals within long user contexts, such as movie titles, user ratings, and paper titles.
\subsubsection{Examples of Summarization Results}
\label{Sec:Summarization Resukts}
We show some summarized results under different tokens for MovieLens and Lamp-5 dataset in Figure~\ref{fig:movie_profile_summaries} and Figure~\ref{fig:title_summary}, respectively.
\subsection{Baseline Details}
\label{Sec:baselines}
We compare against the following baselines: \textbf{Truncation-based:} (a) \textit{Truncate}~\cite{everaert2024amazonqac} uses the most recent $m$ tokens from user contexts without additional processing. \textbf{Direct summarization:} (b) \textit{Direct Summary}~\cite{zhang2024generative} inputs raw contexts into $\Phi_{\text{Sum}}$ to generate compressed profiles. \textbf{Reasoning-enhanced:} (c) \textit{CoT}~\cite{wei2022chain} prompts $\Phi_{\text{Sum}}$ to think step-by-step for summarization; (d) \textit{Self-Reflection}~\cite{ji2023towards} prompts $\Phi_{\text{Sum}}$ to self-reflect and refine its summary. \textbf{Alternative marking methods:} (e) \textit{Mark All} marks all sentences as important; (f) \textit{Random Mark} randomly marks the same number of sentences as $H^*$; (g) \textit{Prompt-GS} first prompts $\Phi_{\text{Sum}}$ to identify important sentences, marks them to derive $H^*$, then summarizes based on the marked context.

\begin{itemize}
    \item \textit{Truncate}: Directly truncates user contexts to the most recent $m$ tokens without additional processing.
    \item \textit{Direct Summary}: Inputs raw contexts into $\Phi_{\text{Sum}}$ to generate compressed profiles. We evaluate various prompts and select the best-performing one. The prompts are shown in Figure~\ref{fig:movie_direct_summary} for MovieLens and Figure~\ref{fig:lamp5_direct_summary} for LaMP-5.
    \item \textit{CoT}: Prompts $\Phi_{\text{Sum}}$ to think step-by-step for summarization. The prompts are shown in Figure~\ref{fig:movie_cot} and Figure~\ref{fig:lamp5_cot}.
    \item \textit{Self-Reflection}: Prompts $\Phi_{\text{Sum}}$ to self-reflect and refine its summary. The prompts are shown in Figure~\ref{fig:movie_reflection} and Figure~\ref{fig:lamp5_reflection}.
    \item \textit{Random Mark} and \textit{Mark All}: Random Mark randomly selects the same number of sentences as $H^*$ for marking, while Mark All marks all sentences as important. Both use the same summarization prompts as Attn-GS, shown in Figure~\ref{fig:movie_prompt_template} and Figure~\ref{fig:academic_title_prompt}.
    \item \textit{Prompt-GS}: First prompts $\Phi_{\text{Sum}}$ to identify important sentences, marks them to derive $H^*$, then summarizes based on the marked context. The identification prompts are shown in Figure~\ref{fig:prompt_gs_identification} , while the summarization prompts are the same as Attn-GS (Figure~\ref{fig:movie_prompt_template} and Figure~\ref{fig:academic_title_prompt}).
\end{itemize}

\begin{figure*}[!t]
\vspace{-10 pt}
\centering
\resizebox{0.9\textwidth}{!}{
\begin{tcolorbox}[mybox={Examples of Marked Context (MovieLens)}]
\textbf{User basic info: Gender is M, Age is 18-24, Occupation is academic/educator, Total Movies watched is 39, Average Rating is 3.24 out of 5.0.}

\textbf{Movie 1 title: The movie title is Twelve Monkeys.} \textbf{Movie 1 year: Released in 1995.} Movie 1 genres: Genres are Drama, Sci-Fi. Movie 1 summary: A time traveler from a post-apocalyptic future is sent back in time to gather information about a deadly virus that has wiped out most of humanity. \textbf{Movie 1 rating: User gave it 5 stars.} \textbf{Movie 1 rating time: Rated on 2000-12-06 at 18:21:10.}

\textbf{Movie 2 title: The movie title is Good Will Hunting.} Movie 2 year: Released in 1997. \textbf{Movie 2 genres: Genres are Drama.} Movie 2 summary: A brilliant but troubled janitor discovers his mathematical genius with the help of a professor who challenges him to confront his past and embrace his potential. Movie 2 rating: User gave it 5 stars. Movie 2 rating time: Rated on 2000-12-06 at 18:21:03.

\textbf{Movie 3 title: The movie title is Run Lola Run (Lola rennt).} Movie 3 year: Released in 1998. Movie 3 genres: Genres are Action, Crime, Romance. Movie 3 summary: A high-stakes race against time unfolds as Lola must find a way to save her boyfriend's life in just 20 minutes. Movie 3 rating: User gave it 5 stars. Movie 3 rating time: Rated on 2000-12-06 at 18:21:03.

\textbf{Movie 4 title: The movie title is Trainspotting.} Movie 4 year: Released in 1996. Movie 4 genres: Genres are Drama. Movie 4 summary: A group of heroin addicts navigate the gritty streets of Edinburgh, Scotland, in a darkly humorous tale of addiction, friendship, and self-destruction...................Movie 37 rating: User gave it 1 stars. Movie 37 rating time: Rated on 2000-12-06 at 18:11:03.

\textbf{Movie 38 title: The movie title is Star Trek VI: The Undiscovered Country.} Movie 38 year: Released in 1991. Movie 38 genres: Genres are Action, Adventure, Sci-Fi. Movie 38 summary: A thrilling and suspenseful adventure as the crew of the USS Enterprise must prevent an interstellar war and uncover a conspiracy threatening peace in the galaxy. Movie 38 rating: User gave it 1 stars. Movie 38 rating time: Rated on 2000-12-06 at 18:11:03.
\end{tcolorbox}}

\caption{Examples of marked context on the MovieLens dataset.The bold texts are marked as important signals (marked between <START\_IMPORTANT> and <END\_IMPORTANT> tags)}
\label{fig:movielens_context}

\end{figure*}

\begin{figure*}[!t]

\centering
\resizebox{0.9\textwidth}{!}{
\begin{tcolorbox}[mybox={Examples of Marked Context (Lamp5)}]
\textbf{Paper 1 title: Visualizing time-oriented data-A systematic view.} Paper 1 abstract: The analysis of time-oriented data is an important task in many application scenarios. In recent years, a variety of techniques for visualizing such data have been published. This variety makes it difficult for prospective users to select methods or tools that are useful for their particular task at hand. In this article, we develop and discuss a systematic view on the diversity of methods for visualizing time-oriented data. With the proposed categorization we try to untangle the visualization of time-oriented data, which is such an important concern in Visual Analytics. \textbf{The categorization is not only helpful for users, but also for researchers to identify future tasks in Visual Analytics..}

\textbf{Paper 1 date: Published in 2007. Paper 2 title: Visualizing Statistical Properties of Smoothly Brushed Data Subsets.} Paper 2 abstract: In many application fields, the statistical properties of data sets are of great interest for data analysts. Since local variations can occur especially in large datasets, it is useful to visualize not only global values,........

\end{tcolorbox}}

\caption{Examples of marked context on the Lamp-5 dataset.The bold texts are marked as important signals (marked between <START\_IMPORTANT> and <END\_IMPORTANT> tags)}
\label{fig:lamp5_context}
\vspace{-10 pt}
\end{figure*} 

\begin{figure*}[htpb]
\vspace{-10 pt}
\centering
\resizebox{0.9\textwidth}{!}{
\begin{tcolorbox}[mybox={Summarization Prompt for MovieLens}]
\small
The information marked between \texttt{<START\_IMPORTANT>} and \texttt{<END\_IMPORTANT>} tags contains critical personalization signals. Prioritize these details in your summary as they are essential for understanding user preferences and behavior patterns.

Generate a \{args.max\_tokens\}-token first-person summary that:
\begin{enumerate}
    \item Prioritizes all important-marked information 
    \item Uses first-person perspective (\texttt{"I am..."}, \texttt{"I like..."})
    \item Preserves key details when possible (names, dates, ratings, etc.)
    \item Output ONLY the summary directly without introductory phrases
    \item Be concise to maximize information density
\end{enumerate}

\end{tcolorbox}}

\caption{Attention-guided summarization prompt template for MovieLens.}
\label{fig:movie_prompt_template}

\end{figure*}

\begin{figure*}[]
\vspace{-10 pt}
\centering
\resizebox{0.9\textwidth}{!}{
\begin{tcolorbox}[mybox={Summarization Prompt for Academic Title Generation}]
The information marked between \texttt{<START\_IMPORTANT>} and \texttt{<END\_IMPORTANT>} tags contains critical personalization signals. Prioritize these details in your summary as they are essential for understanding user preferences and behavior patterns.

Generate a \{args.max\_tokens\}-token first-person summary that:
\begin{enumerate}
    \item Prioritizes all important-marked information
    \item About the users' title writting patterns. optimized for the subsequent academic title generation.
    \item Preserves key details when possible
    \item Output ONLY the summary directly without introductory phrases
    \item Be concise to maximize information density
\end{enumerate}

\end{tcolorbox}}

\caption{Attention-guided summarization prompt template for academic title generation.}
\label{fig:academic_title_prompt}

\end{figure*}

\begin{figure*}[!t]
\vspace{-10 pt}
\centering
\resizebox{0.9\textwidth}{!}{
\begin{tcolorbox}[mybox={Examples of User Profile and Movie Preferences}]
\textbf{50 tokens}
I am an 18-24 year old female doctor who has watched 25 movies with a 4.15 average rating. I love romantic comedies like Notting Hill, You've Got Mail, and Four Weddings and a Funeral, giving most 4-5 stars. I particularly enjoyed Hope Floats, Bed of Roses, and 101 Dalmatians with perfect 5-star ratings, and rated all my movies on July 2, 2000.

\textbf{100 tokens}
I am an 18-24 year old female doctor who has watched 25 movies with a 4.15 average rating. I love romantic comedies and gave 5 stars to classics like Notting Hill, You've Got Mail, Hope Floats, Bed of Roses, E.T., and 101 Dalmatians. I consistently rated popular rom-coms like Forrest Gump, Runaway Bride, Princess Bride, Never Been Kissed, My Best Friend's Wedding, and Four Weddings and a Funeral with 4 stars. I rated all these movies on July 2, 2000, showing my preference for heartwarming stories with romantic elements, comedic adventures, and feel-good narratives from the 1990s era.

\textbf{150 tokens}
I am an 18-24 year old female doctor who has watched 25 movies with an average rating of 4.15 stars. I rated most of my movies on July 2, 2000, showing a clear preference for romantic comedies and dramas. I gave 5-star ratings to 101 Dalmatians, E.T., Bed of Roses, Notting Hill, and Hope Floats. My favorites include romantic films like Forrest Gump, Runaway Bride, You've Got Mail, My Best Friend's Wedding, Four Weddings and a Funeral, and Ghost. I particularly enjoy stories about love, relationships, and emotional journeys. I rated Shakespeare in Love and So I Married an Axe Murderer lower at 3 stars. My viewing patterns suggest I gravitate toward heartwarming narratives with romantic elements, comedic situations, and character-driven plots that explore human connections and personal growth.

\textbf{200 tokens}
I am a female healthcare professional, aged 18-24, who has watched 25 movies with an average rating of 4.15 out of 5 stars. I have a strong preference for romantic comedies and dramas, consistently rating films like Notting Hill, You've Got Mail, and Bed of Roses with 5 stars. My viewing pattern shows I rated all my movies on July 2, 2000, suggesting I discovered cinema around that time. I particularly enjoy heartwarming stories like Forrest Gump, romantic tales like Four Weddings and a Funeral and My Best Friend's Wedding, and even gave Hope Floats a perfect 5-star rating. I also appreciate family-friendly adventures like Flipper and E.T., rating both highly. My taste spans from classic romances like Ghost and The English Patient to quirky comedies like The Nutty Professor and Never Been Kissed. Even unconventional romantic stories like Chasing Amy earned 4 stars from me. The only film I rated below 4 stars was Shakespeare in Love with 3 stars, along with the thriller-comedy So I Married an Axe Murderer, suggesting I prefer straightforward romantic narratives over complex period pieces or dark humor.

\end{tcolorbox}}

\caption{Examples of attention-guided summaries of MovieLens dataset}
\label{fig:movie_profile_summaries}

\end{figure*}

\begin{figure*}[!t]
\vspace{-10 pt}
\centering
\resizebox{0.9\textwidth}{!}{
\begin{tcolorbox}[mybox={Examples of Title Writing Patterns}]
\textbf{50 tokens:}
I write technically precise titles with systematic structure: [Technical Qualifier] + [Core Method] + [Application Domain] + [Performance Goal]. I favor compound constructions using hyphens, emphasize optimization language ("efficient," "robust," "minimizing"), and target systems/computing domains. My patterns prioritize immediate domain identification and methodological contribution clarity.

\textbf{100 tokens:}
I structure titles with technical qualifiers followed by core methods and application domains. My pattern: [Technical Descriptor] + [System/Algorithm] + [Specific Domain] + [Performance Goal]. I favor compound constructions using hyphens and colons for information density. Semantic clusters include energy efficiency, real-time systems, and distributed computing. I use performance-oriented language ("robust," "efficient," "optimal") with domain-specific acronyms (WCRT, GPU, NFV). Titles progress from methodological contribution to application context, maintaining academic formality while ensuring immediate domain recognition for computer science audiences.

\textbf{150 tokens:}
I structure titles with consistent technical qualifiers followed by core methods and specific application domains. My dominant pattern: [Technical Descriptor] + [System/Algorithm] + [Domain Context] + [Performance Objective]. I favor compound constructions using hyphens and colons to maximize information density, as in "Energy-Efficient Resource Utilization for Heterogeneous Embedded Computing Systems." My semantic preferences cluster around systems optimization, particularly energy efficiency, real-time computing, and distributed architectures. I consistently use performance-oriented language ("robust," "efficient," "optimal," "minimizing") paired with specific technical domains (cloud computing, embedded systems, automotive networks). Strategic acronym usage incorporates established abbreviations (WCRT, CAN, GPU, NFV) for precision. Titles typically progress from broad methodological contribution to specific application context, maintaining academic formality while ensuring immediate domain identification. I prioritize scannable hierarchical information that communicates scope, technical contribution, and target application within computer science and engineering contexts.

\textbf{200 tokens:}
I structure titles using a systematic formula: [Technical Qualifier] + [Core Method/System] + [Specific Application Domain] + [Performance Objective]. My preference for compound constructions with hyphens and colons creates hierarchical information density, as in "Energy-Efficient Resource Utilization for Heterogeneous Embedded Computing Systems." My semantic patterns cluster around systems optimization, particularly energy efficiency, real-time computing, and distributed architectures. I consistently employ performance-oriented language ("robust," "efficient," "optimal," "minimizing") paired with specific technical domains (cloud computing, embedded systems, automotive networks, GPU computing). Technical descriptors typically open my titles: "Robust," "Adaptive," "Parallel," "Energy-Efficient," immediately establishing domain expertise. I strategically incorporate established acronyms (WCRT, CAN, GPU, NFV, IoT) when they enhance precision without sacrificing clarity. My title progression follows a logical flow from broad methodological contribution to specific application context, ensuring immediate domain identification. I favor multi-layered technical specifications that communicate both theoretical contribution and practical application scope. The structure maintains academic formality while maximizing information density for computer science and engineering conference audiences, emphasizing algorithmic innovation within constrained system environments.

\end{tcolorbox}}

\caption{Examples of attention-guided summaries of Lamp-5 dataset}
\label{fig:title_summary}

\end{figure*}

\begin{figure*}[htpb]
\vspace{-10 pt}
\centering
\resizebox{0.9\textwidth}{!}{
\begin{tcolorbox}[mybox={Direct Summarization Prompt for MovieLens}]
\small
Generate a \{args.max\_tokens\}-token first-person summary of the user's movie preferences that:
\begin{enumerate}
    \item Captures the user's viewing history and rating patterns
    \item Uses first-person perspective (\texttt{"I am..."}, \texttt{"I like..."})
    \item Preserves key details when possible (names, dates, ratings, etc.)
    \item Output ONLY the summary directly without introductory phrases
    \item Be concise to maximize information density
\end{enumerate}
\end{tcolorbox}}
\caption{Direct summarization prompt template for MovieLens.}
\label{fig:movie_direct_summary}
\end{figure*}

\begin{figure*}[htpb]
\vspace{-10 pt}
\centering
\resizebox{0.9\textwidth}{!}{
\begin{tcolorbox}[mybox={Direct Summarization Prompt for Academic Title Generation}]
Generate a \{args.max\_tokens\}-token first-person summary that:
\begin{enumerate}
    \item Captures the user's title writing patterns and research interests
    \item Optimized for subsequent academic title generation
    \item Preserves key details when possible
    \item Output ONLY the summary directly without introductory phrases
    \item Be concise to maximize information density
\end{enumerate}
\end{tcolorbox}}
\caption{Direct summarization prompt template for academic title generation.}
\label{fig:lamp5_direct_summary}
\end{figure*}

\begin{figure*}[htbp]
\vspace{-10 pt}
\centering
\resizebox{0.9\textwidth}{!}{
\begin{tcolorbox}[mybox={CoT Summarization Prompt for MovieLens}]
\small
Let's think step by step to summarize the user's movie preferences.

Generate a \{args.max\_tokens\}-token first-person summary that:
\begin{enumerate}
    \item Captures the user's viewing history and rating patterns
    \item Uses first-person perspective (\texttt{"I am..."}, \texttt{"I like..."})
    \item Preserves key details when possible (names, dates, ratings, etc.)
    \item Output ONLY the summary directly without introductory phrases
    \item Be concise to maximize information density
\end{enumerate}
\end{tcolorbox}}
\caption{CoT summarization prompt template for MovieLens.}
\label{fig:movie_cot}
\end{figure*}

\begin{figure*}[htbp]
\vspace{-10 pt}
\centering
\resizebox{0.9\textwidth}{!}{
\begin{tcolorbox}[mybox={CoT Summarization Prompt for Academic Title Generation}]
Let's think step by step to summarize the user's title writing patterns.

Generate a \{args.max\_tokens\}-token first-person summary that:
\begin{enumerate}
    \item Captures the user's title writing patterns and research interests
    \item Optimized for subsequent academic title generation
    \item Preserves key details when possible
    \item Output ONLY the summary directly without introductory phrases
    \item Be concise to maximize information density
\end{enumerate}
\end{tcolorbox}}
\caption{CoT summarization prompt template for academic title generation.}
\label{fig:lamp5_cot}
\end{figure*}

\begin{figure*}[htbp]
\vspace{-10 pt}
\centering
\resizebox{0.9\textwidth}{!}{
\begin{tcolorbox}[mybox={Self-Reflection Summarization Prompt for MovieLens (Step 1: Initial Summary)}]
\small
Generate a \{args.max\_tokens\}-token first-person summary of the user's movie preferences that:
\begin{enumerate}
    \item Captures the user's viewing history and rating patterns
    \item Uses first-person perspective (\texttt{"I am..."}, \texttt{"I like..."})
    \item Preserves key details when possible (names, dates, ratings, etc.)
    \item Be concise to maximize information density
\end{enumerate}
\end{tcolorbox}}

\vspace{5pt}

\resizebox{0.9\textwidth}{!}{
\begin{tcolorbox}[mybox={Self-Reflection Summarization Prompt for MovieLens (Step 2: Refinement)}]
\small
Here is an initial summary of the user's movie preferences:
\{initial\_summary\}

Please reflect on this summary and refine it. Consider:
- Does it capture the most important viewing patterns?
- Are key preferences clearly stated?
- Is any critical information missing?
- Can it be more concise while retaining essential details?

Generate a refined \{args.max\_tokens\}-token first-person summary that:
\begin{enumerate}
    \item Uses first-person perspective (\texttt{"I am..."}, \texttt{"I like..."})
    \item Preserves key details when possible (names, dates, ratings, etc.)
    \item Output ONLY the summary directly without introductory phrases
    \item Be concise to maximize information density
\end{enumerate}
\end{tcolorbox}}
\caption{Self-reflection summarization prompt template for MovieLens (two-step process).}
\label{fig:movie_reflection}
\end{figure*}

\begin{figure*}[htbp]
\vspace{-10 pt}
\centering
\resizebox{0.9\textwidth}{!}{
\begin{tcolorbox}[mybox={Self-Reflection Summarization Prompt for Academic Title Generation (Step 1: Initial Summary)}]
Generate a \{args.max\_tokens\}-token first-person summary that:
\begin{enumerate}
    \item Captures the user's title writing patterns and research interests
    \item Optimized for subsequent academic title generation
    \item Preserves key details when possible
    \item Be concise to maximize information density
\end{enumerate}
\end{tcolorbox}}

\vspace{5pt}

\resizebox{0.9\textwidth}{!}{
\begin{tcolorbox}[mybox={Self-Reflection Summarization Prompt for Academic Title Generation (Step 2: Refinement)}]
Here is an initial summary of the user's title writing patterns:
\{initial\_summary\}

Please reflect on this summary and refine it. Consider:
- Does it capture key writing patterns and preferences?
- Are recurring themes and terminology clearly identified?
- Is any critical stylistic information missing?
- Can it be more concise while retaining essential details?

Generate a refined \{args.max\_tokens\}-token first-person summary that:
\begin{enumerate}
    \item Captures the user's title writing patterns optimized for subsequent academic title generation
    \item Preserves key details when possible
    \item Output ONLY the summary directly without introductory phrases
    \item Be concise to maximize information density
\end{enumerate}
\end{tcolorbox}}
\caption{Self-reflection summarization prompt template for academic title generation (two-step process).}
\label{fig:lamp5_reflection}
\end{figure*}

\begin{figure*}[htbp]
\vspace{-10 pt}
\centering
\resizebox{0.9\textwidth}{!}{
\begin{tcolorbox}[mybox={Prompt-GS Identification Prompt}]
\small
You are given a user's personalization context and a task description. Your task is to identify and extract the most important sentences that are relevant for the given task.

Please read through the context carefully and identify sentences that contain critical personalization signals. Copy and paste only the important sentences to the output, preserving their original text exactly.

\textbf{Input format:}

Context: \{context\}

Task Description: \{task\_description\}

\textbf{Output format:} Simply list the important sentences, one per line, without any additional explanation or formatting.
\end{tcolorbox}}
\caption{Prompt-GS identification prompt for extracting important sentences.}
\label{fig:prompt_gs_identification}
\end{figure*}

\label{sec:appendix}

\end{document}